\PassOptionsToPackage{table,dvipsnames}{xcolor}

\documentclass[11pt]{article}
\usepackage{amssymb}
\usepackage{array}

\usepackage[preprint]{acl}

\usepackage{times}
\usepackage{latexsym}

\usepackage[T1]{fontenc}

\usepackage[utf8]{inputenc}

\usepackage{microtype}

\usepackage{inconsolata}

\usepackage{graphicx}
\usepackage{multirow}
\usepackage{booktabs}
\definecolor{darkgreen}{RGB}{0, 128, 0}
\usepackage{amsmath}
\usepackage{enumitem}
\usepackage{mathtools}

\newcommand\Myperm[2][^n]{\prescript{#1\mkern-2.5mu}{}P_{#2}}
\newcommand\Mycomb[2][^n]{\prescript{#1\mkern-0.5mu}{}C_{#2}}

\title{\textbf{\textit{MyCulture}}: Exploring Malaysia's Diverse Culture under \\ Low-Resource Language Constraints}

\author{
Zhong Ken Hew \quad Jia Xin Low \quad Sze Jue Yang \quad Chee Seng Chan\\
Universiti Malaya \\
}

\begin{document}
\maketitle
\begin{abstract}
Large Language Models (LLMs) often exhibit cultural biases due to training data dominated by high-resource languages like English and Chinese. This poses challenges for accurately representing and evaluating diverse cultural contexts, particularly in low-resource language settings. To address this, we introduce \textit{\textbf{MyCulture}}, a benchmark designed to comprehensively evaluate LLMs on Malaysian culture across six pillars: arts, attire, customs, entertainment, food, and religion presented in Bahasa Melayu. Unlike conventional benchmarks, \textit{\textbf{MyCulture}} employs a novel \textit{open-ended multiple-choice question} format without predefined options, thereby reducing guessing and mitigating format bias. We provide a theoretical justification for the effectiveness of this open-ended structure in improving both fairness and discriminative power. Furthermore, we analyze structural bias by comparing model performance on structured versus free-form outputs, and assess language bias through multilingual prompt variations. Our evaluation across a range of regional and international LLMs reveals significant disparities in cultural comprehension, highlighting the urgent need for culturally grounded and linguistically inclusive benchmarks in the development and assessment of LLMs.

\end{abstract}

\section{Introduction}
Large Language Models (LLMs), trained on trillions of tokens exhibit strong generalization and problem-solving abilities beyond simple memorization. Fundamentally, these models are trained on massive text corpora scraped from a vast array of Internet sources.
However, since English and Chinese content dominate the web \cite{w3techs_languages}, LLMs trained on such data are inherently biased toward these high-resource languages, both in terms of linguistic capability and cultural representation \cite{wang2024countriescelebratethanksgivingcultural}.

\begin{figure}[t]
    \centering
    \includegraphics[width=0.9\columnwidth]{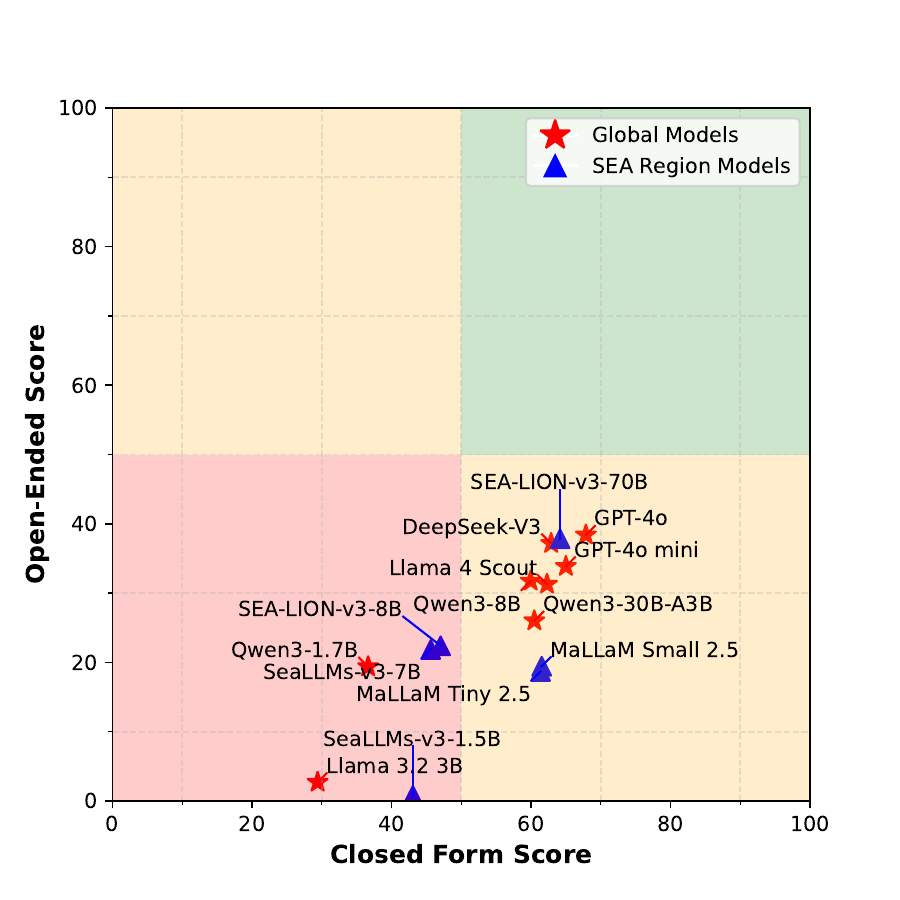}
    \caption{Visualization of the LLMs' scores across both closed-form and open-ended MCQs. The top-right quadrant denotes the optimal region, as ideally, LLMs should be performant in both forms of MCQs.}
    \label{fig:pareto}
\end{figure}

In contrast, resource-scarce languages are often underrepresented, which limits the effectiveness of LLMs in regions where such languages are prevalent. This can result in culturally detached responses that are misaligned with local values and practices, potentially exacerbating issues of social exclusion and racial insensitivity \cite{li2024culturellm}. Ideally, LLMs should exhibit cultural alignment, particularly in a multilingual, multi-ethnic nation such as Malaysia, where inclusivity is constitutionally and culturally emphasized across diverse communities, including Malays, Chinese, Indians, and indigenous groups, as well as multiple religions such as Islam, Buddhism, and Hinduism.

To evaluate cultural alignment, recent efforts such as Kulture \cite{wang2024kulture}, IndoCulture \cite{koto2024indoculture}, SaudiCulture \cite{ayash2025saudiculture} have proposed country-specific cultural benchmarks.
While effective, these benchmarks tend to focus on relatively homogeneous cultural contexts, and thus fall short in addressing the complexities of multi-ethnic, multi-religious societies. A culturally inclusive benchmark that captures this diversity remains an unmet need.

At the same time, as LLMs' capabilities continue to evolve, conventional multiple-choice question (MCQ) formats have shown limitations in accurately capturing models' performance, especially in real-world scenarios that demand free-form generation \cite{li2024multiplechoicequestionsreallyuseful}. While MCQs are straightforward to construct and evaluate, their structured nature tends to favor LLMs by significantly narrowing the combinatorial search space. This increases the likelihood of models arriving atthe correct answer through pattern recognition or random guessing, rather than true comprehension.

In contrast, open-ended questions are better in reflecting real-world use cases, requiring the model to generate responses without being guided by predefined options. However, such formats often involve complex data collection and evaluation pipelines, sometimes necessitating human-in-the-loop assessment which makes them less scalable and less commonly adopted in benchmark design.

Motivated to bridge cultural gaps in a multi-racial and multi-ethnic society, we propose \textit{\textbf{MyCulture}}, an expert-validated, locally contextualized benchmark designed to evaluate LLMs’ understanding and alignment with Malaysian cultural knowledge. \textit{\textbf{MyCulture}} employs a novel open-ended multiple-choice question format without predefined options, thereby reducing guessing and mitigating format bias. We provide a theoretical justification for the effectiveness of this open-ended structure in improving both fairness and discriminative power. The benchmark features carefully crafted questions across diverse cultural domains to assess both factual accuracy and cultural sensitivity in model responses.

Our contributions are as follows:

\begin{itemize}
    \item We introduce \textbf{\textit{MyCulture}}, the first culture benchmark for Malaysia, that is (i) presented in Malay (Bahasa Malaysia), (ii) grounded with locally contextualized questions, and (iii) validated by human experts. It consists of over \emph{2,600} questions, covering six domains across multiple races, religions and ethnicities.

    \item We propose a novel \emph{open-ended} MCQ structure, composed of (i) multi-answer, (ii) ordering and (iii) matching question types, which significantly reduces the probability of LLMs answering correctly via random guessing, ensuring thorough assessment of LLMs' understanding towards local culture.

    \item We showcase the empirical effectiveness of \textbf{\textit{MyCulture}} via extensive experiments. 
    Notably, by simply shifting from closed-form to open-ended MCQs, LLMs' performance deteriorates at least 17\%, signifying an illusion of LLMs grasping cultural knowledge well enough in Malaysia's context.
\end{itemize}

\section{Related Works}
\textbf{Cultural Benchmarks.} 
Assessing LLMs’ cultural understanding is an ongoing effort, as prior work \cite{wang2024countriescelebratethanksgivingcultural} demonstrates that LLMs are culturally biased partly due to the inherent nature of training corpora being dominated by web-scraped content in high-resource languages and cultures.
With these dominant cultures embedded, LLMs tend to reflect dominant cultural perspectives, which oftentimes, neglect local norms and sensitivities.
In light of understanding the state of LLMs cultural understanding towards a nation's context (e.g., Korea), Kulture \cite{wang2024kulture} and KorNAT \cite{lee2024kornat} constructed Korean-native benchmarks from textbooks and surveys.
Meanwhile, J-MMMU \cite{onohara2024jmmmu} reflects LLMs' understanding of Japanese culture by combining both culture-agnostic and culture-specific tasks, outlining that LLMs' generic capabilities in a wide range of tasks do not equivalently transfer to Japanese cultural reasoning.

Beyond single-nation benchmarks, several efforts have focused on evaluating LLMs’ cultural knowledge across a broader international scope. For instance, \citet{myung2024blend} proposed a multilingual benchmark spanning 16 countries or regions, comprising 52.6k manually curated question–answer pairs in 13 languages. Similarly, \citet{chiu2025culturalbenchrobustdiversechallenging} introduced a culturally diverse dataset containing 1,227 human-authored and verified questions covering 45 global regions, including underrepresented areas and 17 topical domains, in both single- and multiple-answer formats.
These works, although impactful, are frequently multinational in scope but monocultural at the national level. Such evaluations would in turn, evaluate LLMs' cultural understanding in a siloed manner, without considering the joint mixture of multiple cultures within a single nation.

\noindent\textbf{Evaluation methods of LLMs.}
Generally, to effectively assess LLMs capabilities, researchers proposed to evaluate LLMs via closed-loop systems such as MCQs.
MCQs are characterized by \emph{fixed options} and \emph{objective scoring}, which are highly feasible in terms of preparing and scoring.
Notable examples of such benchmarks are MMLU \cite{hendrycks2021measuringmassivemultitasklanguage} , CMMLU \cite{li2023cmmlu}, MalayMMLU \cite{poh2024malaymmlu}.
However, with finite options, MCQs inevitably creates a false impression of LLMs are performing well, as LLMs have higher chances of randomly guessing the correct answer.
In contrast, open-ended evaluations are promising direction of reflecting LLMs capabilities in real world use cases, as evident by Arena-Hard Auto \cite{arenahard2024} and IFEval \cite{zhou2023instructionfollowingevaluationlargelanguage}
However, it is usually laborious to collect such a dataset and evaluating LLMs against it poses challenges, as the ground truth answer typically requires human-in-the-loop verification.

Motivated by limitations of existing works, we present a multi-ethnic, multi-racial cultural benchmark for Malaysia, composed of open-ended MCQs.

\section{Theoretical Foundations of Open-Ended MCQ Evaluation}
\label{sec:theory}
In this section, we discuss our theoretical foundation on the proposed \emph{open-ended MCQs}.

\subsection{Evaluation Hardness of Open-Ended MCQs for LLMs}

We provide a formal justification that open-ended multiple-answer MCQs impose significantly stricter evaluation constraints than conventional MCQs, particularly when evaluating cultural knowledge with language models. Our goal is to establish that only models with true coverage over the complete gold answer set can reliably succeed under this evaluation protocol.

\subsection{Notation and Setup}

Let $\mathcal{A}$ be a finite answer alphabet (e.g., token strings, phrase IDs).  
Each question $q$ is associated with a gold answer set $S^\star \subset \mathcal{A}$ of size $k$ and $m$ number of options, i.e., $|S^\star| = k \geq 2$.  
The model receives a textual prompt $x = \text{prompt}(q)$ and produces an output string which is post-processed into a set of $k$ distinct items:
\[
\hat{S} = g\big(f_\theta(x)\big) \subseteq \mathcal{A},
\]
where $f_\theta$ is the language model with parameters $\theta$, and $g(\cdot)$ is a deterministic extraction function (e.g., rule-based span extraction).

We define the \textbf{exact-set accuracy} as:
\[
\text{Acc}(f_\theta; q) =
\begin{cases}
1, & \text{if } \hat{S} = S^\star, \\
0, & \text{otherwise}.
\end{cases}
\]

\subsection{Baseline: Single answer MCQs}
For conventional four-option MCQs, even a random-guessing model $f_{\text{rand}}$ achieves:
\[
P_{\text{MCQ}} = \frac{1}{4}, \text{ where } P_\text{MCQ} = \frac{1}{m} \text{, } m=4
\]
where $m$ refers to number of options. This reflects the format’s vulnerability to guessing and surface-level pattern recognition.

\subsection{Open-Ended MCQ: Factorial Decay in Guessability}

We now consider the case where the model outputs $k$ answers from $\mathcal{A}$, chosen uniformly at random without replacement. Then the probability of selecting exactly the gold answer set is:

\begin{equation}
\Pr\big(\hat{S} = S^\star\big) = \frac{1}{\binom{n}{k}}, \quad \text{where } n = |\mathcal{A}|.
\label{eq:openmcqprob}
\end{equation}

\paragraph{Lemma 1 (Factorial Decay).}
If $n = k$ (i.e., only the $k$ gold answers exist), then:
\[
P_{\text{open}} = \frac{1}{k!}.
\]

\textit{Proof.} In this case, $\binom{k}{k} = k!$, so the model has only one correct permutation out of $k!$ total sets. 

\paragraph{Example Values.}
\begin{itemize}
  \item For $k = 4$: $P_{\text{open}} = \frac{1}{24} \approx 4.17\%$
  \item For $k = 6$: $P_{\text{open}} = \frac{1}{720} \approx 0.14\%$
\end{itemize}

\paragraph{Lemma 2 (Combinatorial Gap).}
Let $P_{\text{MCQ}} = 1/m$ and $P_{\text{open}} = 1/\binom{n}{k}$, then:
\[
\frac{P_{\text{open}}}{P_{\text{MCQ}}} = \frac{m}{\binom{n}{k}} \leq \frac{m}{k!},
\]
so the open-ended format is at least $k!/m$ times harder than MCQ under random guessing.

\subsection{Information-Theoretic Analysis}

We now examine the Shannon information conveyed by each question format. Let $I$ denote the surprisal (information content) of a correct guess.

\begin{align*}
I_{\text{open}} &= -\log_2 \left( \frac{1}{\binom{n}{k}} \right) = \log_2 \binom{n}{k}, \\
I_{\text{MCQ}}  &= \log_2 m.
\end{align*}

\paragraph{Example:} For $n = 8$ and $k = 4$:
\[
\binom{8}{4} = 70, \quad \Rightarrow \quad I_{\text{open}} \approx \log_2 70 \approx 6.13 \text{ bits}, \]

$\quad I_{\text{MCQ}} = \log_2 4 = 2 \text{ bits}.$

Thus, each open-ended MCQ provides over 3× more evaluative signal than a standard MCQ.

\subsection{Link to Autoregressive LLMs}

Let the language model be autoregressive:
\[
p_\theta(\hat{S} \mid x) = \prod_{i=1}^{k} p_\theta(a_i \mid x, a_1, \dots, a_{i-1}),
\]
where $a_i \in \hat{S}$.
If the model fails to assign significant probability mass to even a single culturally correct answer $a_j \in S^\star$, then:
\[
p_\theta(\hat{S} = S^\star \mid x) \approx 0,
\]
pushing the expected accuracy toward the hard bound of Equation~\ref{eq:openmcqprob}.  
This shows that only LLMs with full coverage of the gold cultural space can consistently succeed under this scoring scheme.

\begin{figure}[h!]
  \centering
  \includegraphics[width=1\columnwidth]{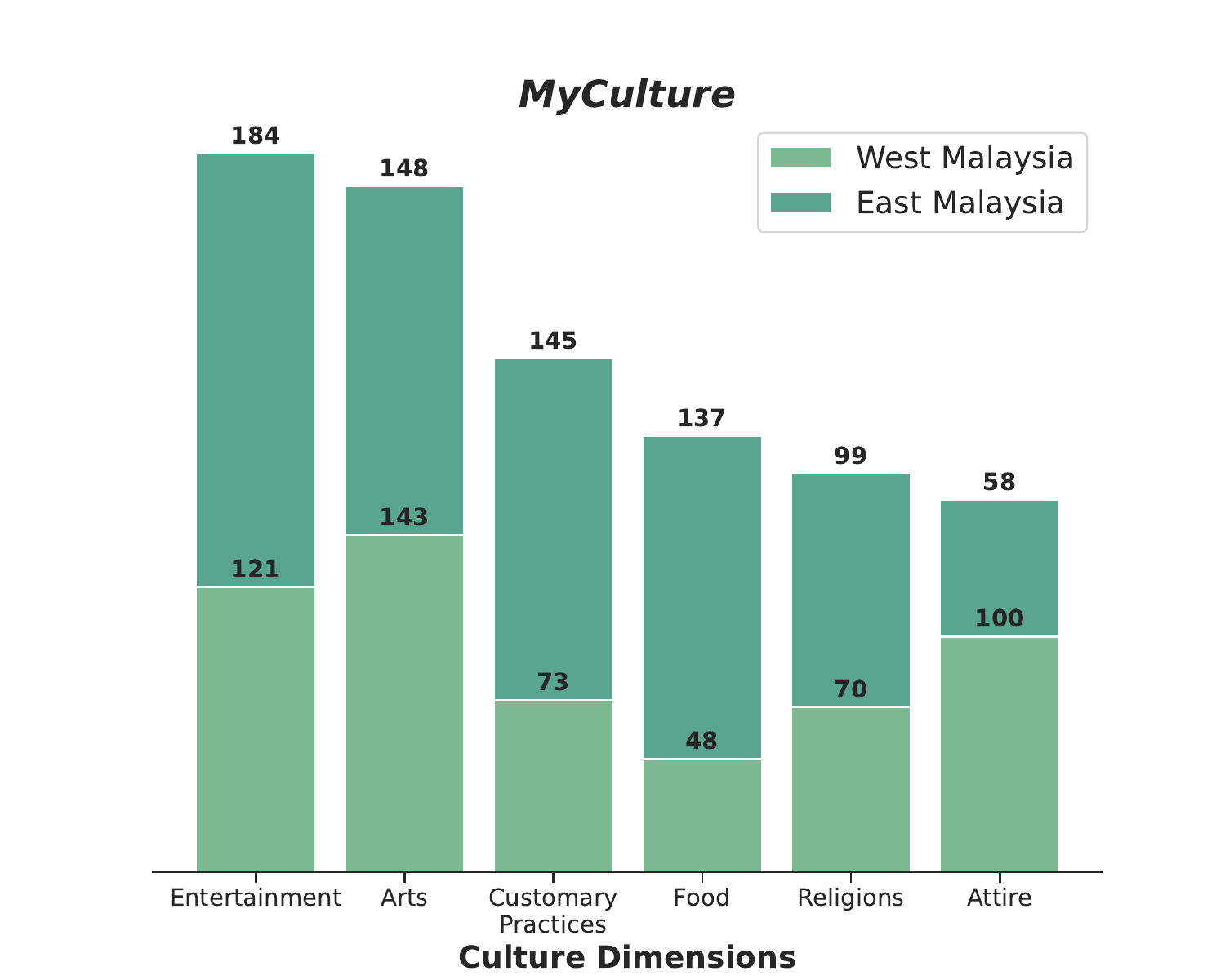}
  \caption{Regional data distribution of \textbf{\textit{MyCulture}} in a single format, across East and West Malaysia. The figure illustrates the number of culturally grounded questions curated for each region, highlighting differences in representation across the six cultural pillars. This distribution reflects both regional diversity and the outcome of quality filtering during dataset construction.}
  \label{fig:mycul_stat}
\end{figure}

\section{\textit{\textbf{MyCulture}}}
In this section, we formally introduce \textbf{\textit{MyCulture}}, a Malaysia-based cultural benchmark dataset, which covers six core dimensions, namely: (i) \emph{arts}, (ii) \emph{attire}, (iii) \emph{customary practices}, (iv) \emph{entertainment}, (v) \emph{food} and finally, (vi) \emph{religions}.
\textbf{\textit{MyCulture}} is carefully designed to include diverse races, ethnicities and geographical locations within Malaysia, to accurately reflect local norms.
Additionally, it incorporates the proposed \emph{open-ended MCQ} format, with three variants: (i) \emph{multi-answer}, (ii) \emph{ordering} and (iii) \emph{matching}.
These variants, ranked from easiest to hardest, from \emph{multi-answer} to \emph{matching}, with a decreasing probability of randomly guessing the correct answer.



\subsection{Dataset Statistics}
\noindent\textbf{Distribution.}
\textbf{\textit{MyCulture}} consists of a total of 2,652 questions, which are equally divided into 1,326 closed-form and open-ended MCQs, respectively.
Each set of the MCQs (closed-form/open-ended) consists of questions from six main cultural dimensions across East and West Malaysia, where the distributions are depicted in Figure~\ref{fig:mycul_stat}.
Notably, both the closed-form and open-ended MCQs are rigorously validated by local experts, ensuring the quality and validity of the questions.

\noindent\textbf{Language.}
The MCQs are constructed in Malay (known as \emph{Bahasa Malaysia}), which is the national language of Malaysia.
Since Malay is markedly a low-resource language, construction of the dataset involves extensive human validation to ensure its correctness.




\subsection{Question Structure}
\label{sec:question-structure}
\noindent\textbf{Multi-Answer (MA) MCQs.}
In a conventional closed-form manner, MCQs contain multi-answer questions that present $N$ statements (e.g., I, II, III) for LLMs to select from, and typically, these questions offer \textbf{one} correct answer that consists of all correct statements.
We further transform such MA MCQs into their open-ended counterparts, by encoding the statements as the answer choices (see Table~\ref{tab:malaysia_mcq_examples} for a comparison).
With such a transformation to open-ended MA MCQs, we reduced the probability of random guessing to:


\begin{equation}
  P_{\text{MA}}\bigl(\mathcal{A}\bigr) = \frac{1}{\displaystyle\sum_{r=1}^{m} \Mycomb[m]{r}}; \text{ where } m = N
  \label{eq:mamcq}
\end{equation}
where $r$ denotes the number of choices to be selected from a pool of $m$ options, and \(P_{\text{MA}}(\mathcal{A})\) $<$ $P_\text{MCQ}(\mathcal{A})$.

\noindent\textbf{Ordering (O) MCQs}
are a set of MCQs that requires LLMs to rearrange a sequence of $N$ statements into a chronological order and in a closed-form manner, they present $m$ choices with a predefined fixed orders, similar to normal MCQs.
The open-ended form of O-MCQs requires LLMs to provide all of the correct statements, in an ``exact'' manner.
The probability of guessing the correct sequence correctly is formulated as:

\begin{equation}
  P_{\text{O}}\bigl(\mathcal{A}\bigr) = \frac{1}{\displaystyle\sum_{r=1}^{m} \Myperm[m]{r}}; \text{ where } m = N
  \label{eq:omcq}
\end{equation}
Such a formulation guarantees $P_\text{O}(\mathcal{A}) < P_\text{MCQ}(\mathcal{A})$.

\noindent\textbf{Matching (M) MCQs}
are tasked evaluate LLMs' capabilities in associating $N$ items from two sets of options.
For closed-form formats, M-MCQs possess no difference compared to conventional MCQs; however, in the open-ended form, to accurately answer such MCQs, LLMs are required to generate the correct pair by matching options from the two sets.
The probability of getting the correct answer is as follows:

\begin{equation}
P_{\text{M}}\bigl(\mathcal{A}\bigr) = \frac{1}{\prod_{n=1}^{m} n^2}; \text{ where } m=N, \quad m > 1
\label{eq:mmcq}
\end{equation}
where \(P_{\text{M}}(\mathcal{A}) < P_\text{MCQ}(\mathcal{A})\).

\begin{figure}[t]
  \centering
  \includegraphics[width=\columnwidth]{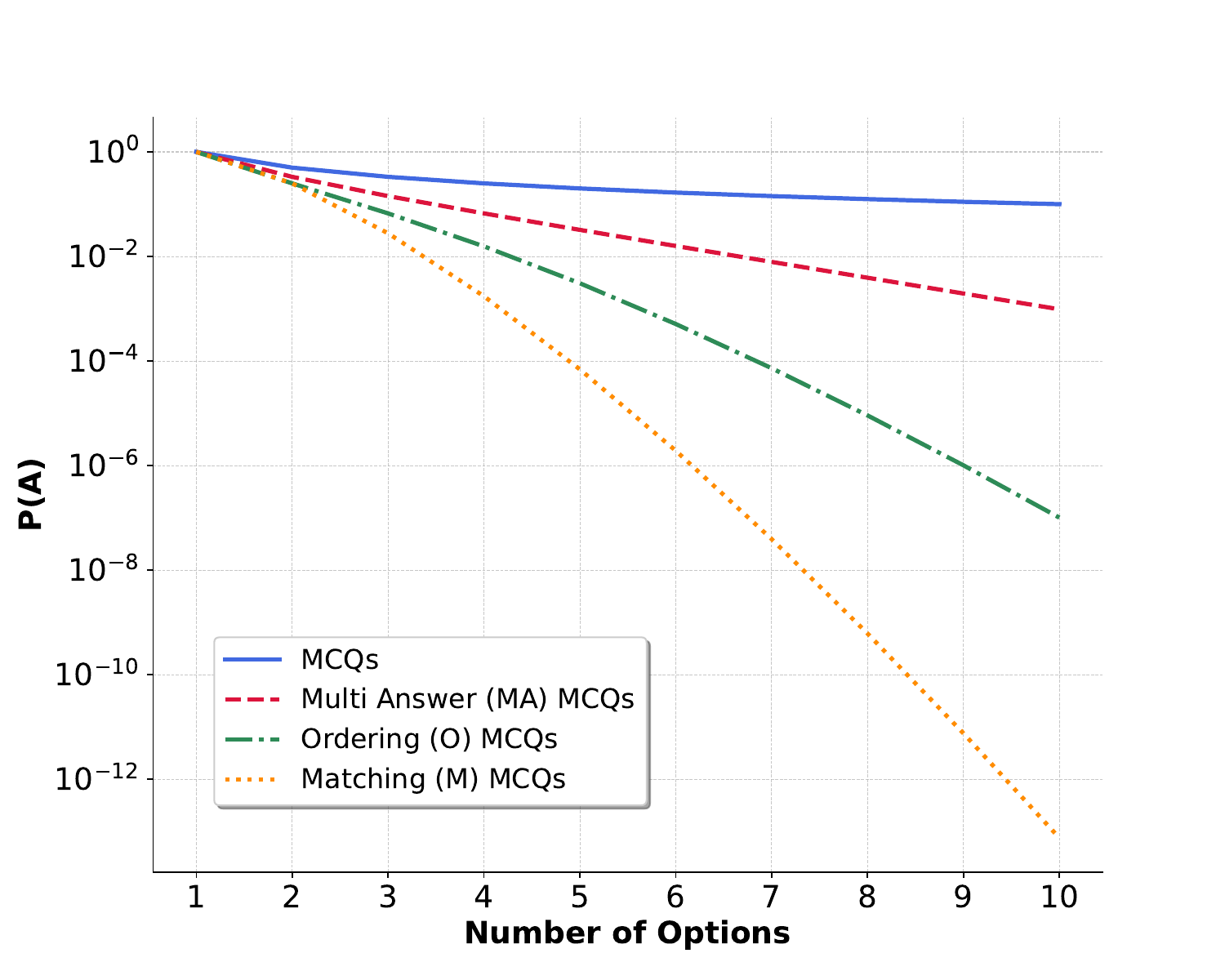}
  \caption{Visualization of the probability of correctly answering by random guessing under the closed-form format and the proposed open-ended format.}
  \label{fig:prob_formula}
\end{figure}

These three formats are designed to expand the effective combinatorial search space inherent to open-ended MCQs, thereby substantially reducing the probability of correct responses through random guessing. 
Figure~\ref{fig:prob_formula} presents a comparative visualization of the probability of selecting the correct answer under the proposed formulations versus the closed-form MCQs  format with predefined options. The x-axis denotes the number of options (for closed-form MCQs ) or the number of statements (for open-ended MCQs formats), demonstrating that the probability of random correctness decreases as this number increases. Notably, the decline in probability is markedly steeper for the three proposed formats than for the closed-form MCQs  format, indicating a substantial increase in task difficulty. This formulation ensures the following probabilistic ordering:

\[
P_{\text{M}}(\mathcal{A}) \;<\;
P_{\text{O}}(\mathcal{A}) \;<\;
P_{\text{MA}}(\mathcal{A})\;<\;
P_{\text{MCQ}}(\mathcal{A})
\]
As evident in Figure~\ref{fig:prob_formula}, a typical 4-option MCQ will yield $P(\mathcal{A})=0.25$, while $P_\text{MA}(\mathcal{A})=0.0667$, $P_\text{MA}(\mathcal{A})=0.0156$ and $P_\text{M}(\mathcal{A})=0.0017$, respectively, signifying an increased difficulty in guessing the correct answer.
Such a design mitigates format-induced biases and facilitates a more rigorous and discriminative evaluation of model capabilities across diverse reasoning tasks, specifically for cultural understanding.

\section{Experiment \& Analysis}
In this section, we rigorously evaluate LLMs' performance against \textbf{\textit{MyCulture}}, via extensive empirical experiments.

\subsection{Experiment Setting}
We evaluate a total of \textbf{14} LLMs under zero-shot settings, which includes latest closed-source models (GPT-4o, GPT-4o mini) \cite{gpt-4}, open-source models such as DeepSeek-V3 \cite{deepseek}, Qwen3 \cite{qwen3}, Llama \cite{llama} and LLMs trained on predominantly on Southeast Asia (SEA) data which include SEA-LLM \cite{seallm}, SEA-LION \cite{sealion}, MaLLaM \cite{mallam}.
All of the models are evaluated against both closed-form and open-ended MCQs, and we outline the results in Table~\ref{tab:llm_performance}.

\begin{table*}[h!]
\centering
\footnotesize
\setlength{\tabcolsep}{5pt}
\renewcommand{\arraystretch}{1.2}
\begin{tabular}{p{0.48\textwidth}|p{0.48\textwidth}}
\hline
\centering\textbf{\textit{Closed-form MCQs}} & \centering\textbf{\textit{Open-ended MCQs}} \tabularnewline
\hline
\multicolumn{2}{c}{\textbf{Multiple-answer MCQs}} \tabularnewline
\hline
\raggedright
\textbf{Question:} Yang manakah antara berikut berkaitan dengan makanan tradisional Malaysia? \newline
\textbf{Pernyataan:} \newline
I. Nasi lemak biasanya dihidang dengan sambal dan telur. \newline
II. Satay berasal dari Thailand. \newline
III. Rendang sering disediakan semasa perayaan. \newline
IV. Kuih-muih tradisional hanya popular di Sabah. \newline
\textbf{Pilihan:} \newline 
A. I, III \newline 
B. I, II, III \newline
C. II, IV \newline 
D. I, III, IV \newline
\textbf{Answer:} A
&
\raggedright
\textbf{Question:} Yang manakah berkaitan dengan makanan tradisional Malaysia? \newline
A. Nasi lemak dihidang dengan sambal dan telur \newline
B. Satay berasal dari Thailand \newline
C. Rendang disediakan semasa perayaan \newline
D. Kuih tradisional hanya popular di Sabah \newline
\textbf{Answer:} A, C
\tabularnewline
\hline
\multicolumn{2}{c}{\textbf{Matching MCQs}} \tabularnewline
\hline
\raggedright
\textbf{Question:} Padankan pakaian tradisional dengan negeri asal. \newline
\textbf{Kumpulan A:} \newline
A. Baju Melayu Cekak Musang \newline
B. Baju Kurung Teluk Belanga \newline
C. Busana Iban \newline
D. Baju Bodo \newline
\textbf{Kumpulan B:} \newline
i. Johor \newline
ii. Sarawak \newline
iii. Sabah \newline
iv. Terengganu \newline
\textbf{Pilihan:} \newline 
A: A-iv, B-i, C-ii, D-iii \newline 
B: A-iv, B-ii, C-i, D-iii \newline 
C: A-iv, B-iii, C-i, D-ii \newline 
D: A-iv, B-ii, C-iii, D-i \newline 
\textbf{Answer:} A
&
\raggedright
\textbf{Question:} Padankan pakaian tradisional dengan negeri asal. \newline
A. Baju Melayu Cekak Musang \newline
B. Baju Kurung Teluk Belanga \newline
C. Busana Iban \newline
D. Baju Bodo \newline
i. Terengganu \newline
ii. Johor \newline
iii. Sarawak \newline
iv. Sabah \newline
\textbf{Answer:} A-i, B-ii, C-iii, D-iv
\tabularnewline
\hline
\multicolumn{2}{c}{\textbf{Ordering MCQs}} \tabularnewline
\hline
\raggedright
\textbf{Question:} Susun aktiviti utama dalam sambutan Tahun Baru Cina di Malaysia. \newline
\textbf{Pernyataan:} \newline
I. Mengemas rumah \newline
II. Makan besar bersama keluarga \newline
III. Memberi angpau \newline
IV. Persembahan tarian singa \newline
\textbf{Pilihan:} A. I, II, III, IV \quad B. II, I, III, IV \quad C. I, III, II, IV \quad D. I, II, IV, III \newline
\textbf{Answer:} A
&
\raggedright
\textbf{Question:} Susun aktiviti utama dalam sambutan Tahun Baru Cina di Malaysia. \newline
A. Mengemas rumah \newline
B. Makan besar bersama keluarga \newline
C. Memberi angpau \newline
D. Persembahan tarian singa \newline
\textbf{Answer:} A, B, C, D
\tabularnewline
\hline
\end{tabular}
\caption{Examples of open-ended MCQs reformulated from their closed-form counterparts. The contents of the MCQs are similar across both forms, however, in open-ended MCQs, instead of selecting a single correct answer, LLMs are required to provide the exact answer, for \emph{multi-answer}, \emph{matching} and \emph{ordering} MCQs.}
\label{tab:malaysia_mcq_examples}
\end{table*}

\subsection{Experiment Results}

\begin{table*}[h!]
\centering
\small
\setlength{\tabcolsep}{5pt}
\renewcommand{\arraystretch}{1.1}
\begin{tabular}{@{}lllcccc@{}}
\toprule
\textbf{Company} & \textbf{Models} & \textbf{\# Param} & \textbf{Source} & \textbf{Closed-form MCQs} & \textbf{Open-ended MCQs} & \textbf{Diff} \\
\midrule
\multirow{2}{*}{Open-AI} & GPT-4o & - & Close & \textbf{67.87} & \textbf{38.39} & \textcolor{red}{-29.48} \\
                         & GPT-4o mini & - & Close & 65.00 & 33.86 & \textcolor{red}{-31.14} \\
\midrule
Deepseek & Deepseek-V3* & 37B (671B) & Open & 62.89 & 37.17 & \textcolor{red}{-25.72} \\
\midrule
\multirow{3}{*}{Qwen} & Qwen3* & 3.3B (30B) & Open & 60.48 & 26.01 & \textcolor{red}{-34.47} \\
                      & Qwen3 & 8B & Open & 59.95 & 31.74 & \textcolor{red}{-28.21} \\
                      & Qwen3 & 1.7B & Open & 36.65 & 19.38 & \textcolor{red}{-17.27} \\
\midrule
\multirow{2}{*}{META} & Llama 4 - Scout* & 17B (109B) & Open & 62.30 & 31.28 & \textcolor{red}{-31.02} \\
                      & Llama-3.2 & 3B & Open & 29.41 & 2.71 & \textcolor{red}{-26.70} \\
\midrule
\multicolumn{7}{c}{\textbf{Southeast Asian Region LLM}} \\
\midrule
\multirow{2}{*}{SEALLM} & SeaLLMs-v3 & 7B & Open & 45.63 & 21.87 & \textcolor{red}{-23.76} \\
                        & SeaLLMs-v3 & 1.5B & Open & 43.06 & 0.83 & \textcolor{red}{-42.23} \\
\midrule
\multirow{2}{*}{SEA-Lion} & SEA-LION-V3 & 70B & Open & \underline{64.17} & \underline{37.85} & \textcolor{red}{-26.32} \\
                          & SEA-LION-V3& 8B & Open & 47.05 & 22.40 & \textcolor{red}{-24.65} \\
\midrule
\multirow{2}{*}{Mesolitica} & MaLLaM Small 2.5 & - & Close & 61.53 & 19.45 & \textcolor{red}{-42.08} \\
                            & MaLLaM Tiny 2.5 & - & Close & 61.38 & 18.70 & \textcolor{red}{-42.68} \\
\bottomrule
\end{tabular}
\caption{Performance of LLMs on closed-form MCQs and open-ended MCQs. * indicates Mixture of Expert (MoE) models. \textbf{Bold} indicates the highest performance across all models. \underline{Underline} indicates the highest performance among Southeast Asian regional LLMs. \textbf{Diff} indicates the performance differences between closed-form and open-ended MCQs. Best viewed in color.}
\label{tab:llm_performance}
\end{table*}

We reported zero-shot results of 14 LLMs on 2 formats of \textbf{\textit{MyCulture}}, closed-form and open-ended MCQs, in Table~\ref{tab:llm_performance}.
In Figure~\ref{fig:pareto}, we visualize the performance of these LLMs, where the top-right quadrant denotes the ideal performance.

\noindent\textbf{Closed-form MCQs.} 
Based on Table~\ref{tab:llm_performance}, the GPT series consistently achieves the highest overall performance, with GPT-4o and GPT-4o mini ranking first and second, respectively. Among open-source models, SEA-LION (70B) attains the best performance, suggesting that fine-tuning on Southeast Asian data improves cultural alignment with Malaysian contexts. Additionally, models with over 100B parameters (Deepseek-V3; Llama-4 Scout) exhibit comparable performance, scoring around 62\%. In the sub-10B category, Qwen3 (8B) outperforms its peers, demonstrating competitive cultural understanding despite its smaller size.
Within Southeast Asian regional models, SEA-LION (70B) leads in performance, likely due to its larger parameter count. Notably, MaLLaM, a Malaysia-centric LLM achieves performance on par with prominent international models such as DeepSeek-V3, Qwen3 (30B-A3B), and Llama 4 Scout, highlighting its effectiveness in capturing Malaysia-specific cultural knowledge.

\noindent\textbf{Open-ended MCQs.}
In open-ended MCQ settings, we observe a similar performance trend among LLMs, with GPT-4o consistently achieving the best results and SEA-LION (70B) emerging as the top performer among open-source models. Despite this consistency, transitioning from closed-form to open-ended MCQs causes a substantial performance decline across all evaluated models. As illustrated in Table \ref{tab:llm_performance}, this degradation varies widely among models, ranging from a maximum drop of 42.68 points for MaLLaM Tiny (2.5B) to a minimum of 17.27 points for Qwen3 (1.7B). These results indicate that relying solely on closed-form MCQs may create an illusion of higher model competence. The absence of explicit answer options in open-ended MCQs demands that LLMs not only interpret the cultural context embedded within each question but also independently generate accurate responses. This requirement necessitates a nuanced and thorough understanding of Malaysian culture, significantly increasing the complexity of the task. Therefore, open-ended MCQs constitute a more rigorous and insightful evaluation paradigm and should be adopted in future assessments of LLMs' capabilities.

\noindent\textbf{Insights.}
As shown in Figure~\ref{fig:pareto}, most models perform better on the closed-form version of \textit{\textbf{MyCulture}}, as majority of them score above the 50\% mark. 
In contrast, in open-ended MCQs with identical contents, LLMs consistently have sub-par performance compared to their closed-form counterparts ($<$50\%). 

\subsection{Ablation Study}
\begin{figure}[t]
  \centering
  \includegraphics[width=1\columnwidth]{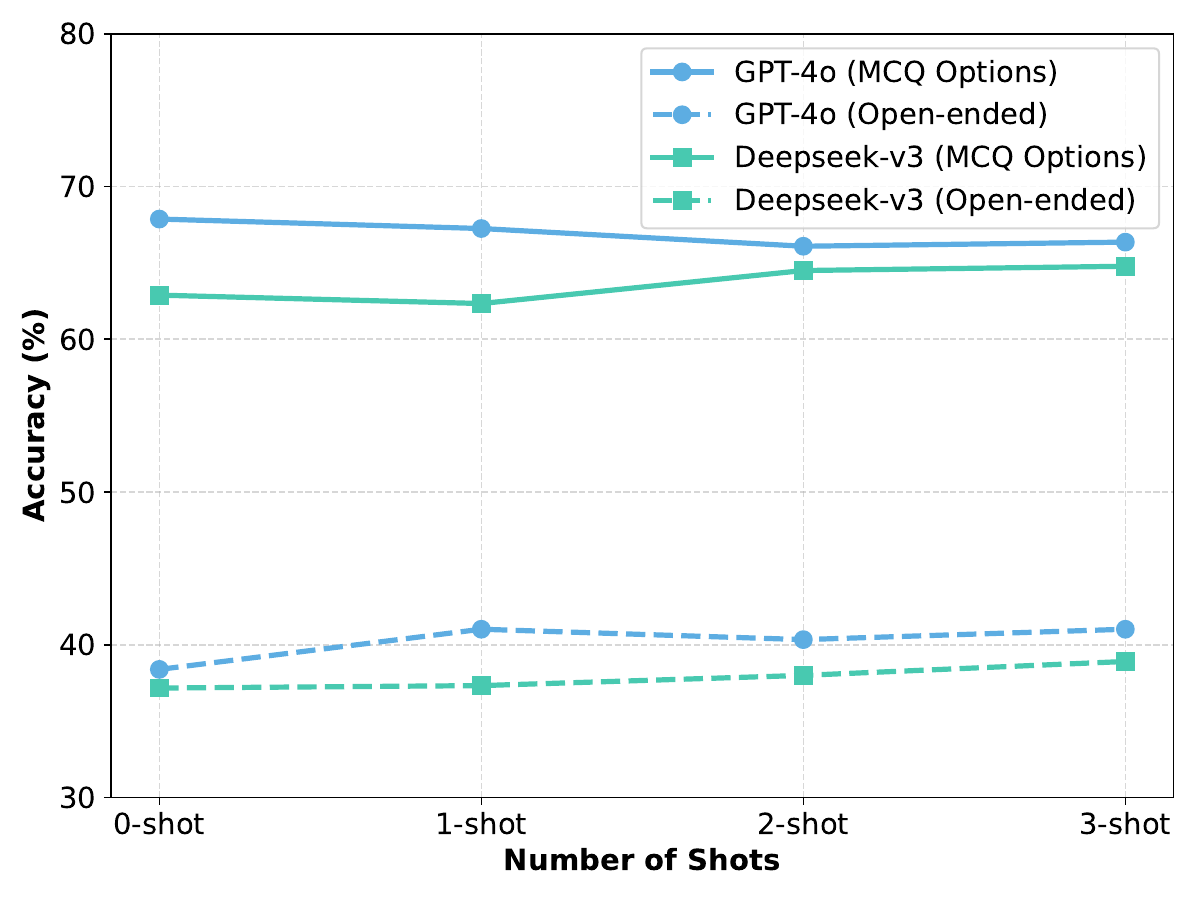}
  \caption{Comparison of model performance under normal and proposed formats across different shot settings. Notably, there is no apparent trend of improvements across increasing number of shots.}
  \label{fig:shot_comparison}
\end{figure}

\noindent\textbf{Few-shot examples.}
Following prior work on  benchmarks \cite{poh2024malaymmlu}, we also investigate the effect of few-shot prompting on LLM performance. In the few-shot setting, we randomly sample a number of examples from the \textit{\textbf{MyCulture}} dataset and include them in the system prompt. To ensure a fair evaluation, any question appearing in the prompt is excluded from the test set.
Figure~\ref{fig:shot_comparison} presents the performance of GPT-4o and DeepSeek-V3 across varying shot configurations. The results show minimal variation in performance across different shot counts, consistent with previous findings \cite{li2023cmmlu} that few-shot prompting provides limited benefit in cultural understanding tasks.

\begin{table*}[h!]
\centering
\small
\setlength{\tabcolsep}{5pt}
\renewcommand{\arraystretch}{1.15}
\begin{tabular}{@{}lllccc@{}}
\toprule
\textbf{Company} & \textbf{Model} & \textbf{\# Param} & \textbf{Unstructured} & \textbf{Structured} & \textbf{Diff} \\
\midrule
\multirow{2}{*}{OpenAI} 
  & GPT-4o & - & 38.39 & 41.03 & \textcolor{darkgreen}{+2.64} \\
  & GPT-4o mini & - & 33.86 & 37.40 & \textcolor{darkgreen}{+3.54} \\
\midrule
\rowcolor{gray!30}
Deepseek 
  & Deepseek-V3* & 37B (671B) & 37.17 & 2.87 & \textcolor{red}{-34.30} \\
\midrule
\multirow{3}{*}{Qwen}
  & Qwen3* & 3.3B (30B) & 26.01 & 30.24 & \textcolor{darkgreen}{+4.23} \\
  & Qwen3 & 8B & 31.74 & 27.98 & \textcolor{red}{-3.76} \\
  & Qwen3 & 1.7B & 19.38 & 18.10 & \textcolor{red}{-1.28} \\
\midrule
\multirow{2}{*}{Meta}
  & Llama 4 - Scout* & 17B (109B) & 31.28 & 29.71 & \textcolor{red}{-1.57} \\
  & Llama-3.2 & 3B & 2.71 & 17.94 & \textcolor{darkgreen}{+15.23} \\
\midrule
\multicolumn{6}{c}{\textbf{Southeast Asian Regional LLMs}} \\
\midrule
\multirow{2}{*}{SEALLM}
  & SeaLLMs-v3 & 7B & 21.87 & 12.90 & \textcolor{red}{-8.97} \\
  & SeaLLMs-v3 & 1.5B & 0.83 & 8.14 & \textcolor{darkgreen}{+7.31} \\
\midrule
\multirow{2}{*}{SEA-LION}
  & SEA-LION-V3 & 70B & 37.85 & \textbf{43.06} & \textcolor{darkgreen}{+5.21} \\
  & SEA-LION-V3 & 8B & 22.40 & 26.92 & \textcolor{darkgreen}{+4.52} \\
\midrule
\multirow{2}{*}{Mesolitica}
  & MaLLaM Small 2.5 & - & 19.45 & 38.31 & \textcolor{darkgreen}{+18.86} \\
  & MaLLaM Tiny 2.5 & - & 18.70 & 38.15 & \textcolor{darkgreen}{+19.45} \\
\bottomrule
\end{tabular}
\caption{
Performance of LLMs on closed-form and open-ended MCQs. * indicates Mixture of Expert (MOE) models. \textbf{Bold} indicates the highest performance across all models. \underline{Underline} indicates the highest performance among Southeast Asian regional LLMs. \textbf{Diff} shows the performance differences between structured and unstructured outputs. DeepSeek’s performance is greyed out as it fails to follow the system prompt for generating structured output, resulting in significantly lower scores. Best viewed in color.}
\label{tab:structure_bias}
\end{table*}

\noindent\textbf{Structured and unstructured outputs.}
\cite{raspanti2025grammar} studies the effect of constraining LLMs with specific structures (i.e., limiting the set of possible output tokens), in terms of grammatical structures.
However, the effect of such constrained decoding is unexplored in cultural nuances.
Hence, we investigate the effect of structured outputs in terms of Malaysia culture, by evaluate LLMs in 2 different output settings, \emph{structured output} and \emph{free form generation} (unstructured output). 
Table \ref{tab:structure_bias} presents the performance of various LLMs under two different output formats. In general, LLMs perform better with structured outputs, suggesting that guiding models to follow a predefined format can enhance their accuracy in answering MCQs. This improvement is likely due to the constraint that helps LLMs focus on generating the intended answer that is complex as show in Table \ref{tab:structure_bias} rather than producing lengthy explanations, which may include hallucinations or omit the answer altogether, which is coherent with prior works \cite{raspanti2025grammar, banerjee2025cranereasoningconstrainedllm}.
However, DeepSeek-V3 failed to adhere to the predefined JSON schema, showing a significant degradation from unstructured to structured outputs. 

\begin{table*}[h!]
\centering
\small
\setlength{\tabcolsep}{6pt}
\renewcommand{\arraystretch}{1.2}
\begin{tabular}{llccc}
\toprule
\textbf{Company} & \textbf{Models} & \textbf{Malay Prompt} & \textbf{English Prompt} & \textbf{Chinese Prompt} \\
\midrule
\multirow{2}{*}{Open-AI}
  & GPT-4o & 38.39 & \textbf{38.98} & 37.93 \\
  & GPT-4o mini & 33.86 & 34.31 & \textbf{34.69} \\
\midrule
Deepseek & Deepseek-v3 & 37.17 & \textbf{40.79} & 39.29 \\
\midrule
\multirow{2}{*}{SEA-Lion}
  & SEA-LION-V3 (70b) & 37.85 & 39.14 & \textbf{39.51} \\
  & SEA-LION-V3 (8b) & \textbf{22.40} & 8.06 & 20.13 \\
\midrule
\multirow{2}{*}{Mesolitica}
  & MaLLam small 2.5 & 19.45 & 23.53 & \textbf{25.56} \\
  & MaLLam tiny 2.5 & 18.70 & 23.60 & \textbf{25.64} \\
\bottomrule
\end{tabular}
\caption{Overall performance across different language system prompts on \textbf{\textit{MyCulture}}. \textbf{Bold} indicates the highest performance of models across different languages of the same question.}
\label{tab:language-per}
\end{table*}

\noindent\textbf{Language biases.}
We examine the impact of prompt language on LLM performance with \textbf{\textit{MyCulture}}. \citet{myung2024blend} shows that LLMs perform better when language and culture are aligned.
However, their study involves cultures that are expressed in a single language, which differ from multi-lingual country like Malaysia, that
oftentimes, express a single culture in multiple languages, which remain underexplored in the domain of LLMs.
Hence, we study the effect of a culture across multiple languages, by augmenting the instructions to LLMs into different local languages, as depicted in Table~\ref{tab:language-per}.
In contrast to the findings of \citet{myung2024blend}, our results suggest that using the local language (Malay) does not lead to improved performance. Instead, prompts in high-resource languages such as English and Chinese yield significantly better results. This discrepancy may stem from the low-resource nature of Malay, while Chinese and English benefit from broader pretraining coverage. Furthermore, the strong performance under Chinese and English prompts implies that LLMs primarily acquire knowledge of Malaysian culture through these languages. 

\section{Conclusion}
In this paper, we introduced \textit{\textbf{MyCulture}}, a culturally grounded benchmark designed to evaluate LLMs’ understanding of Malaysia’s multi-ethnic and multi-religious context. Unlike prior benchmarks that focus on culturally homogeneous settings, \textbf{\textit{MyCulture}} captures the diversity of Malaysian society through 2,600 questions written in Bahasa Malaysia, composed of both closed-form and open-ended multiple-choice formats across six cultural domains. To address limitations of conventional MCQs, which often inflate LLM performance through format-induced bias, we proposed a novel open-ended structure comprising multi-answer, ordering, and matching types without predefined options. Our empirical analysis shows that simply shifting to this open-ended format leads to a $\geq$17\% drop in model accuracy, revealing an overestimation of LLMs’ true cultural comprehension. These findings underscore the importance of culturally contextualized and structurally robust benchmarks for fair and meaningful evaluation of LLM capabilities.

\bibliography{main}

\begin{thebibliography}{25}
\providecommand{\natexlab}[1]{#1}

\bibitem[{Achiam et~al.(2023)Achiam, Adler, Agarwal, Ahmad, Akkaya, Aleman, Almeida, Altenschmidt, Altman, Anadkat et~al.}]{gpt-4}
Josh Achiam, Steven Adler, Sandhini Agarwal, Lama Ahmad, Ilge Akkaya, Florencia~Leoni Aleman, Diogo Almeida, Janko Altenschmidt, Sam Altman, Shyamal Anadkat, and 1 others. 2023.
\newblock Gpt-4 technical report.
\newblock \emph{arXiv preprint arXiv:2303.08774}.

\bibitem[{Ayash et~al.(2025)Ayash, Alhuzali, Alasmari, and Aloufi}]{ayash2025saudiculture}
Lama Ayash, Hassan Alhuzali, Ashwag Alasmari, and Sultan Aloufi. 2025.
\newblock Saudiculture: A benchmark for evaluating large language models’ cultural competence within saudi arabia.
\newblock \emph{Journal of King Saud University Computer and Information Sciences}, 37(6):123.

\bibitem[{Banerjee et~al.(2025)Banerjee, Suresh, Ugare, Misailovic, and Singh}]{banerjee2025cranereasoningconstrainedllm}
Debangshu Banerjee, Tarun Suresh, Shubham Ugare, Sasa Misailovic, and Gagandeep Singh. 2025.
\newblock \href {https://arxiv.org/abs/2502.09061} {Crane: Reasoning with constrained llm generation}.
\newblock \emph{Preprint}, arXiv:2502.09061.

\bibitem[{Chiu et~al.(2025)Chiu, Jiang, Lin, Park, Li, Ravi, Bhatia, Antoniak, Tsvetkov, Shwartz, and Choi}]{chiu2025culturalbenchrobustdiversechallenging}
Yu~Ying Chiu, Liwei Jiang, Bill~Yuchen Lin, Chan~Young Park, Shuyue~Stella Li, Sahithya Ravi, Mehar Bhatia, Maria Antoniak, Yulia Tsvetkov, Vered Shwartz, and Yejin Choi. 2025.
\newblock \href {https://arxiv.org/abs/2410.02677} {Culturalbench: A robust, diverse, and challenging cultural benchmark by human-ai culturalteaming}.
\newblock \emph{Preprint}, arXiv:2410.02677.

\bibitem[{Dubey et~al.(2024)Dubey, Jauhri, Pandey, Kadian, Al-Dahle, Letman, Mathur, Schelten, Yang, Fan et~al.}]{llama}
Abhimanyu Dubey, Abhinav Jauhri, Abhinav Pandey, Abhishek Kadian, Ahmad Al-Dahle, Aiesha Letman, Akhil Mathur, Alan Schelten, Amy Yang, Angela Fan, and 1 others. 2024.
\newblock The llama 3 herd of models.
\newblock \emph{arXiv e-prints}, pages arXiv--2407.

\bibitem[{Hendrycks et~al.(2021)Hendrycks, Burns, Basart, Zou, Mazeika, Song, and Steinhardt}]{hendrycks2021measuringmassivemultitasklanguage}
Dan Hendrycks, Collin Burns, Steven Basart, Andy Zou, Mantas Mazeika, Dawn Song, and Jacob Steinhardt. 2021.
\newblock \href {https://arxiv.org/abs/2009.03300} {Measuring massive multitask language understanding}.
\newblock \emph{Preprint}, arXiv:2009.03300.

\bibitem[{Koto et~al.(2024)Koto, Mahendra, Aisyah, and Baldwin}]{koto2024indoculture}
Fajri Koto, Rahmad Mahendra, Nurul Aisyah, and Timothy Baldwin. 2024.
\newblock Indoculture: Exploring geographically influenced cultural commonsense reasoning across eleven indonesian provinces.
\newblock \emph{Transactions of the Association for Computational Linguistics}, 12:1703--1719.

\bibitem[{Lee et~al.(2024)Lee, Kim, Kim, Kim, Won, Lee, and Choi}]{lee2024kornat}
Jiyoung Lee, Minwoo Kim, Seungho Kim, Junghwan Kim, Seunghyun Won, Hwaran Lee, and Edward Choi. 2024.
\newblock Kornat: Llm alignment benchmark for korean social values and common knowledge.
\newblock \emph{arXiv preprint arXiv:2402.13605}.

\bibitem[{Li et~al.(2024)Li, Chen, Wang, Sitaram, and Xie}]{li2024culturellm}
Cheng Li, Mengzhuo Chen, Jindong Wang, Sunayana Sitaram, and Xing Xie. 2024.
\newblock Culturellm: Incorporating cultural differences into large language models.
\newblock \emph{Advances in Neural Information Processing Systems}, 37:84799--84838.

\bibitem[{Li et~al.(2023)Li, Zhang, Koto, Yang, Zhao, Gong, Duan, and Baldwin}]{li2023cmmlu}
Haonan Li, Yixuan Zhang, Fajri Koto, Yifei Yang, Hai Zhao, Yeyun Gong, Nan Duan, and Timothy Baldwin. 2023.
\newblock Cmmlu: Measuring massive multitask language understanding in chinese.
\newblock \emph{arXiv preprint arXiv:2306.09212}.

\bibitem[{Li* et~al.(2024)Li*, Chiang*, Frick, Dunlap, Zhu, Gonzalez, and Stoica}]{arenahard2024}
Tianle Li*, Wei-Lin Chiang*, Evan Frick, Lisa Dunlap, Banghua Zhu, Joseph~E. Gonzalez, and Ion Stoica. 2024.
\newblock \href {https://lmsys.org/blog/2024-04-19-arena-hard/} {From live data to high-quality benchmarks: The arena-hard pipeline}.

\bibitem[{Li et~al.(2024)Li, Li, Xiang, Liu, Deng, and Garcia}]{li2024multiplechoicequestionsreallyuseful}
Wangyue Li, Liangzhi Li, Tong Xiang, Xiao Liu, Wei Deng, and Noa Garcia. 2024.
\newblock \href {https://arxiv.org/abs/2403.17752} {Can multiple-choice questions really be useful in detecting the abilities of llms?}
\newblock \emph{Preprint}, arXiv:2403.17752.

\bibitem[{Liu et~al.(2024)Liu, Feng, Xue, Wang, Wu, Lu, Zhao, Deng, Zhang, Ruan et~al.}]{deepseek}
Aixin Liu, Bei Feng, Bing Xue, Bingxuan Wang, Bochao Wu, Chengda Lu, Chenggang Zhao, Chengqi Deng, Chenyu Zhang, Chong Ruan, and 1 others. 2024.
\newblock Deepseek-v3 technical report.
\newblock \emph{arXiv preprint arXiv:2412.19437}.

\bibitem[{Myung et~al.(2024)Myung, Lee, Zhou, Jin, Putri, Antypas, Borkakoty, Kim, Perez-Almendros, Ayele et~al.}]{myung2024blend}
Junho Myung, Nayeon Lee, Yi~Zhou, Jiho Jin, Rifki Putri, Dimosthenis Antypas, Hsuvas Borkakoty, Eunsu Kim, Carla Perez-Almendros, Abinew~Ali Ayele, and 1 others. 2024.
\newblock Blend: A benchmark for llms on everyday knowledge in diverse cultures and languages.
\newblock \emph{Advances in Neural Information Processing Systems}, 37:78104--78146.

\bibitem[{Ng et~al.(2025)Ng, Nguyen, Huang, Tai, Leong, Leong, Yong, Ngui, Susanto, Cheng et~al.}]{sealion}
Raymond Ng, Thanh~Ngan Nguyen, Yuli Huang, Ngee~Chia Tai, Wai~Yi Leong, Wei~Qi Leong, Xianbin Yong, Jian~Gang Ngui, Yosephine Susanto, Nicholas Cheng, and 1 others. 2025.
\newblock Sea-lion: Southeast asian languages in one network.
\newblock \emph{arXiv preprint arXiv:2504.05747}.

\bibitem[{Nguyen et~al.(2023)Nguyen, Zhang, Li, Aljunied, Hu, Shen, Chia, Li, Wang, Tan et~al.}]{seallm}
Xuan-Phi Nguyen, Wenxuan Zhang, Xin Li, Mahani Aljunied, Zhiqiang Hu, Chenhui Shen, Yew~Ken Chia, Xingxuan Li, Jianyu Wang, Qingyu Tan, and 1 others. 2023.
\newblock Seallms--large language models for southeast asia.
\newblock \emph{arXiv preprint arXiv:2312.00738}.

\bibitem[{Onohara et~al.(2024)Onohara, Miyai, Imajuku, Egashira, Baek, Yue, Neubig, and Aizawa}]{onohara2024jmmmu}
Shota Onohara, Atsuyuki Miyai, Yuki Imajuku, Kazuki Egashira, Jeonghun Baek, Xiang Yue, Graham Neubig, and Kiyoharu Aizawa. 2024.
\newblock Jmmmu: A japanese massive multi-discipline multimodal understanding benchmark for culture-aware evaluation.
\newblock \emph{arXiv preprint arXiv:2410.17250}.

\bibitem[{Poh et~al.(2024)Poh, Yang, Tan, Chieng, Tan, Yu, Mun, and Chan}]{poh2024malaymmlu}
Soon Poh, Sze~Jue Yang, Jeraelyn Tan, Lawrence Chieng, Jia Tan, Zhenyu Yu, Foong Mun, and Chee~Seng Chan. 2024.
\newblock Malaymmlu: A multitask benchmark for the low-resource malay language.
\newblock In \emph{Findings of the Association for Computational Linguistics: EMNLP 2024}, pages 650--669.

\bibitem[{Raspanti et~al.(2025)Raspanti, Ozcelebi, and Holenderski}]{raspanti2025grammar}
Federico Raspanti, Tanir Ozcelebi, and Mike Holenderski. 2025.
\newblock Grammar-constrained decoding makes large language models better logical parsers.
\newblock In \emph{Proceedings of the 63rd Annual Meeting of the Association for Computational Linguistics (Volume 6: Industry Track)}, pages 485--499.

\bibitem[{{W3Techs}(2025)}]{w3techs_languages}
{W3Techs}. 2025.
\newblock Usage statistics of content languages for websites.
\newblock \url{https://w3techs.com/technologies/overview/content_language}.
\newblock Accessed: 2025-08-07.

\bibitem[{Wang et~al.(2024{\natexlab{a}})Wang, Jiao, Huang, Dai, tse Huang, Tu, and Lyu}]{wang2024countriescelebratethanksgivingcultural}
Wenxuan Wang, Wenxiang Jiao, Jingyuan Huang, Ruyi Dai, Jen tse Huang, Zhaopeng Tu, and Michael~R. Lyu. 2024{\natexlab{a}}.
\newblock \href {https://arxiv.org/abs/2310.12481} {Not all countries celebrate thanksgiving: On the cultural dominance in large language models}.
\newblock \emph{Preprint}, arXiv:2310.12481.

\bibitem[{Wang et~al.(2024{\natexlab{b}})Wang, Yeo, Lim, and Kim}]{wang2024kulture}
Xiaonan Wang, Jinyoung Yeo, Joon-Ho Lim, and Hansaem Kim. 2024{\natexlab{b}}.
\newblock Kulture bench: A benchmark for assessing language model in korean cultural context.
\newblock \emph{arXiv preprint arXiv:2412.07251}.

\bibitem[{Yang et~al.(2025)Yang, Li, Yang, Zhang, Hui, Zheng, Yu, Gao, Huang, Lv et~al.}]{qwen3}
An~Yang, Anfeng Li, Baosong Yang, Beichen Zhang, Binyuan Hui, Bo~Zheng, Bowen Yu, Chang Gao, Chengen Huang, Chenxu Lv, and 1 others. 2025.
\newblock Qwen3 technical report.
\newblock \emph{arXiv preprint arXiv:2505.09388}.

\bibitem[{Zhou et~al.(2023)Zhou, Lu, Mishra, Brahma, Basu, Luan, Zhou, and Hou}]{zhou2023instructionfollowingevaluationlargelanguage}
Jeffrey Zhou, Tianjian Lu, Swaroop Mishra, Siddhartha Brahma, Sujoy Basu, Yi~Luan, Denny Zhou, and Le~Hou. 2023.
\newblock \href {https://arxiv.org/abs/2311.07911} {Instruction-following evaluation for large language models}.
\newblock \emph{Preprint}, arXiv:2311.07911.

\bibitem[{Zolkepli et~al.(2024)Zolkepli, Razak, Adha, and Nazhan}]{mallam}
Husein Zolkepli, Aisyah Razak, Kamarul Adha, and Ariff Nazhan. 2024.
\newblock Mallam--malaysia large language model.
\newblock \emph{arXiv preprint arXiv:2401.14680}.

\end{thebibliography}

\end{document}